\documentclass[11pt]{article}

\usepackage[preprint]{acl}

\usepackage{times}
\usepackage{latexsym}

\usepackage[T1]{fontenc}

\usepackage[utf8]{inputenc}

\usepackage{microtype}

\usepackage{inconsolata}

\usepackage{graphicx}
\usepackage{multirow}

\usepackage{lipsum}
\usepackage{comment}
\usepackage[table,xcdraw]{xcolor}
\usepackage{multirow}
\usepackage{comment}
\usepackage{booktabs}
\usepackage{lipsum}

\usepackage{array}

\usepackage{tcolorbox}
\newcounter{promptno}[section]
\newlength\mystoreparindent
\newenvironment{prompt}[1][]
{
  \setlength{\mystoreparindent}{\the\parindent}
  \setlength{\parindent}{0pt}
  \refstepcounter{promptno}
  \par\medskip
  \noindent
  \begin{tcolorbox}[left=1pt,right=1pt]
  \textsc{{template \small\thesubsection.\thepromptno}}\\
  \small
  \tt
}{
  \end{tcolorbox}
  \setlength{\parindent}{\mystoreparindent}
  \medskip
}

\title{Multi-Turn Multi-Agent Dialogue for Collaborative Reconstruction Improves VLM Performance on Spatial Reasoning, But Only Barely}

\author{%
Chalamalasetti Kranti${^\mathbf{1}}$, Sherzod Hakimov${^\mathbf{1}}$, David Schlangen${^\mathbf{1,2}}$\\$^{\mathbf{1}}$Computational Linguistics, Department of Linguistics\\
University of Potsdam, Germany\\
$^{\mathbf{2}}$German Research Center for Artificial Intelligence (DFKI), Berlin, Germany\\
{\texttt{\{kranti.chalamalasetti, sherzod.hakimov, david.schlangen\}@uni-potsdam.de}}
}

\begin{document}
\maketitle
\begin{abstract}
Robots operating in diverse environments rely on visual input to interpret objects and spatial layouts. In human-collaborative tasks, they are expected to communicate this understanding through language. Vision-language models (VLMs) support robotic tasks involving visual interpretation, question answering, and instruction following, but their capabilities in collaborative dialogue tasks requiring spatial reasoning remain underexplored. We study this gap through a collaborative structure-building task that combines visual interpretation, grounding, language-guided interaction, and action generation. We develop a framework in which VLMs use dialogue to reconstruct a target structure from visual and textual inputs. We evaluate open-weight and closed VLMs across interaction settings, input modalities, and image representations. Results show that spatial reasoning over visual representations remains difficult for the evaluated VLMs. Detailed text representations of the target yield higher reconstruction success across modality conditions, while decomposed image representations improve performance. These findings reveal limits in visual spatial grounding and grounded instruction generation for collaborative VLM agents.
\end{abstract}

\section{Introduction}
\label{sec:introduction}
Many robots deployed in human-robot environments use vision as a primary sensory modality. Their ability to act in these settings depends on interpreting visual input and task-relevant environmental changes, making accurate image interpretation necessary for robot control and effective human-robot interaction. Recent developments in large language models (LLMs) have enabled progress in robotic tasks such as task planning~\citep{DBLP:conf/icra/SinghBMGXTFTG23, DBLP:conf/icra/LiangHXXHIFZ23}, robot behavior guidance~\citep{DBLP:conf/iclr/WangLYSBQWX024, DBLP:conf/iros/HuangLHW024}, and physical grounding~\citep{DBLP:conf/icml/HuangAPM22, pmlr-v205-ichter23a, DBLP:conf/iclr/ZengAICWWTPRSLV23}. Vision-language models (VLMs) further extend this progress to visual perception by supporting vision-language tasks such as image captioning~\citep{DBLP:conf/eccv/ChenLDZHWZL24, DBLP:conf/acl/ChengSFMSXYCZC25} and visual question answering~\citep{DBLP:conf/cvpr/0003XKISGX24}.

\begin{figure}[t]
  \includegraphics[width=\columnwidth]{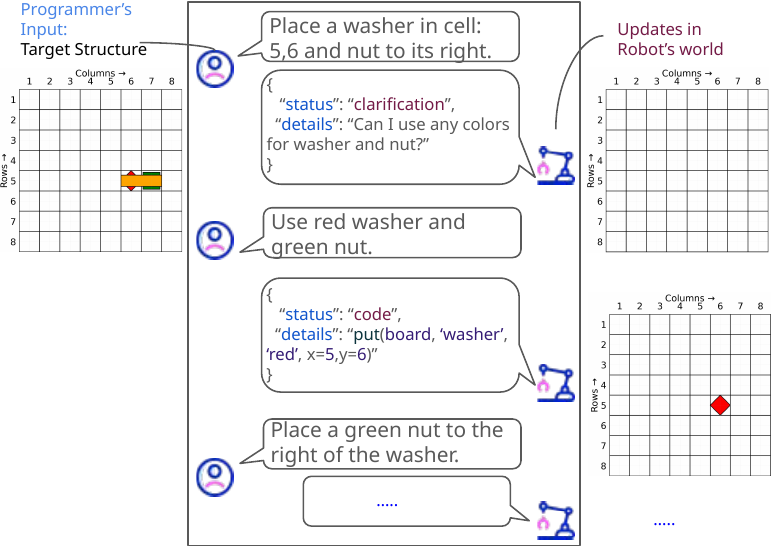}
  \caption{Illustration of the two-agent structure-building task. The Programmer uses the target structure to generate instructions, and the Robot grounds these instructions into executable actions that update its grid state.}
  \label{fig:firstpagefigure}
  \vspace{-0.3cm}
\end{figure}

Despite progress in VLM collaboration~\citep{ji2024foundation, DBLP:conf/rss/ShiHZSPLLF24} and dialogue capabilities~\citep{DBLP:conf/nips/LiuYZYLZL00SZ24, DBLP:conf/cvpr/0001TZC024}, VLMs continue to struggle with spatial reasoning~\citep{DBLP:conf/cvpr/0003XKISGX24, DBLP:conf/emnlp/MayerBJNB25}. However, the source of these failures remains underexplored. Errors may arise from visual perception, language grounding, or misinterpretations that propagate across dialogue turns. Single-turn benchmarks and direct action-generation settings cannot disentangle these sources of failure. To explore these issues in a controlled setting, we propose a structure-building task (see Figure~\ref{fig:firstpagefigure}) in which two VLMs communicate their image interpretations through dialogue.

In this task, the models collaborate to reconstruct a target structure on a grid. One VLM acts as the instruction generator, while the other acts as the instruction translator, guiding object manipulation on the grid. This role separation allows us to systematically analyse whether failures arise from target-structure interpretation, instruction generation, instruction interpretation, or action execution, and how visual and linguistic errors propagate during collaborative structure building.

We instantiate the task using goal structures from an existing grid-based structure-building dataset~\citep{kranti2024towards}. The dataset is suitable for our proposed setting as it contains assembly-style components arranged on $8 \times 8$ 2.5D~\footnote{A 2D grid that also allows vertical stacking of components.} grids with vertical stacking, along with Python code for each structure. This code allows us to render target images that resemble top-down views in assembly settings, while preserving executable structure specifications for evaluation. We evaluate the task in both single-agent and two-agent settings, varying the input modality across text-only, image-only, and text-and-image conditions.

Our contributions are as follows. First, we introduce a controlled structure-building framework that separates the stages of VLM collaboration into target interpretation, instruction generation, instruction interpretation, and action execution. Second, we use this framework to evaluate how interaction setting, target representation, input modality, and structural complexity affect the reconstruction. Third, we analyze dialogue behaviors, including clarification questions and correction instructions, to study how VLM agents respond to uncertainty and execution errors during task completion.

\section{Related Work}

\paragraph{Robotic Manipulation and Structure-Building Tasks}
Several works study tasks that require agents to reason about spatial relations~\citep{4059223, DBLP:conf/naacl/BiskYM16, DBLP:conf/rss/ShridharH18, DBLP:journals/tacl/LachmyPMT22} and action sequences~\citep{DBLP:conf/icra/SinghBMGXTFTG23, DBLP:conf/icra/LiangHXXHIFZ23, kranti2024towards} to manipulate the robot's world~\citep{DBLP:conf/acl/JayannavarNH20, pmlr-v205-ichter23a, jayannavar2026bap}. These works focus on grounding goals into executable actions. Our work follows this line of research, but examines how VLM agents communicate visual and spatial information through dialogue.

\paragraph{VLMs for Grounding and Spatial Reasoning}
Prior work has studied VLMs on multimodal tasks such as image captioning~\citep{DBLP:conf/eccv/ChenLDZHWZL24, DBLP:conf/acl/ChengSFMSXYCZC25}, visual question answering~\citep{DBLP:conf/cvpr/0003XKISGX24, DBLP:journals/access/KimCLR24, DBLP:conf/eccv/KarTPKZT24}, physical grounding~\citep{DBLP:conf/icml/HuangAPM22, pmlr-v205-ichter23a, DBLP:conf/iclr/ZengAICWWTPRSLV23}, and benchmarking~\citep{DBLP:journals/ijcv/KrishnaZGJHKCKL17, DBLP:conf/cvpr/ShridharTGBHMZF20, DBLP:conf/nips/LiuYZYLZL00SZ24}. Despite this progress, VLMs continue to struggle with spatial reasoning tasks~\citep{DBLP:conf/cvpr/0003XKISGX24, DBLP:journals/corr/abs-2503-19707, DBLP:journals/corr/abs-2506-03135, DBLP:conf/emnlp/MayerBJNB25}. However, how VLM performance changes as scene complexity increases remains underexplored.

\paragraph{Collaboration and Dialogue} Dialogue-based tasks have been used to study how agents exchange information and coordinate toward a shared goal. \citet{DBLP:conf/acl/KimKCRZTBP19} proposed \texttt{Co-Draw}, a collaborative drawing task in which one agent describes an abstract scene and another reconstructs it. \citet{DBLP:conf/cvpr/VriesSCPLC17} introduced \texttt{GuessWhat}, a visually grounded dialogue game in which agents identify a target object through questions and answers. \citet{DBLP:conf/aaai/UdagawaA19} studied common grounding under continuous and partially observable contexts, where agents must establish shared understanding through dialogue. Other works evaluate LLMs in self-play and collaborative settings~\citep{DBLP:conf/emnlp/ChalamalasettiG23, chiu-etal-2023-symbolic, DBLP:journals/corr/abs-2308-10032, DBLP:journals/corr/abs-2504-11442}. Our work extends these ideas to structure building task, where VLM agents interpret vision information, communicate spatial information, and reconstruct the target structure over multiple turns.

\section{Methodology}
\label{sec:methodology}
Our objective is to evaluate the spatial reasoning capabilities of VLMs in collaborative dialogue-based structure-building tasks. We formulate the task as a multi-turn multimodal interaction between two players who collaborate to build a target structure.

\subsection{``Self-Play'' for Collaborative Dialogue}
\label{subsec:methodology-vlmselfplay}
Following recent work on self-play interactions~\citep{DBLP:conf/emnlp/ChalamalasettiG23, DBLP:journals/corr/abs-2308-10032, DBLP:journals/corr/abs-2504-11442}, we set up the proposed task as a two-player VLM self-play interaction. The first player (Player1), referred to as the \textit{Programmer}, acts as the instruction giver. The Programmer receives the target structure, the Robot's current grid state, the environment context, and the dialogue history. The second player (Player2), referred to as the \textit{Robot}, acts as the instruction follower. The Robot receives the Programmer's instruction, its current grid state, the environment context, and the dialogue history, but never sees the target structure. The Programmer generates instructions for building the target structure (\textit{e.g., place a yellow washer at the 1st row, 1st column}), while the Robot translates these instructions into executable object-manipulation Python code (\textit{e.g., put(board, `washer', `yellow', row=1, col=1)}); the prompt context indicates the relevant APIs (see Figure~\ref{fig:promptstructure_sagent_part2} in Appendix) to use for object manipulation on the grid. The Robot's response follows a JSON format with two fields: \texttt{status} and \texttt{details}. The \texttt{status} field indicates whether the response contains executable code or a clarification question or acknowledgment, and the \texttt{details} field contains the corresponding content. The prompts for both players (see Figure~\ref{fig:promptstructure_twoagent_prog_mt_1} and Figure~\ref{fig:promptstructure_twoagent_robot_mt}) and ablation studies for the prompt formulation are available in the Appendix~\ref{subsubsec:appendix-prompt-templates}.

We build the two-player interaction based on \texttt{clem:todd}~\citep{kranti-etal-2025-clem}, where the interaction is orchestrated by a Game Master (GM) (see Appendix~\ref{subsec:appendix-interaction-fw} for details). The Game Master receives model responses, validates whether they satisfy the constraints specified in the prompt, cleans the responses and extracting structured fields, and passes the processed outputs between the players. The GM also handles turn-level coordination.

\subsection{Structure Building Task Setup}
\begin{figure}[t]
  \includegraphics[width=0.49\textwidth]{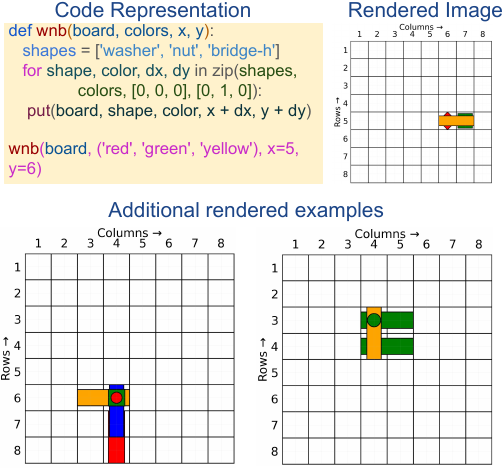}
  \caption{Example of a target structure code and its rendered representation. The top-left panel shows the code representation used to specify component placements, while the top-right panel shows the corresponding rendered grid image. The bottom panels show additional rendered target structures from the dataset.}
  \label{fig:datasetexample}
  \vspace{-0.3cm}  
\end{figure}

We use goal structures from an existing dataset SARTCo~\citep{kranti2024towards}. Each goal structure specifies an arrangement of components on a 2.5D $8 \times 8$ grid, referred to as a \textit{board}. For each board, the dataset includes a target structure specification and corresponding Python code. We execute this code to render the component arrangement as a grid-style image using Matplotlib (see Figure~\ref{fig:datasetexample}). We use the rendered images as the target structures in our proposed task.

The Programmer is prompted with the target structure, the Robot's current state, and the environment information. The interaction proceeds over multiple turns. In the first turn, the Robot's grid is empty, and the Programmer receives the target structure and generates an instruction for building it. The Robot receives the instruction along with its current grid state, which is empty in the first turn. The Robot then either asks a clarification question by indicating the \textit{status} type as \textit{clarification} in its response or generates code (status type as \textit{code}) to update the grid according to the instruction. The GM validates and executes the Robot's generated code in a virtual simulator, and then shares the updated Robot grid state with the Programmer. The Programmer does not observe the Robot's generated code. Instead it observes the updated grid state and uses it to generate the followup instruction, correction, or termination signal. The interaction stops when the Programmer generates \texttt{DONE}, indicating that the Robot grid matches the target structure, or when the interaction reaches the maximum limit of $15$ turns. The interaction is aborted if the Robot-generated code produces execution errors for three consecutive attempts.

\section{Experimental Setup}
\label{sec:expsetup}

\subsection{Dataset} As mentioned in Section~\ref{sec:methodology}, we use SARTCo dataset. The dataset has two styles of boards: simple and regular, for this task we only use simple boards where a single object made by 2--5 elements arranged in a particular style on a $8 \times 8$ grid (see Figure~\ref{fig:datasetexample}). The test set contains \textbf{495 boards in total}: 30 with two elements, 102 with three elements, 219 with four elements, and 144 with five elements. We use this test set for all experiment configurations (more details on dataset are available in Appendix~\ref{subsec:appendix-dataset}).

\subsection{Model Selection} We evaluate two VLMs: an open-weight model, Qwen3-VL-30B-A3B-Instruct, from the Qwen family, and a closed model, GPT-5.2-Chat. We treat this as an illustrative case study, selecting one open-weight and one closed model as representative of two distinct development paradigms. A single A100 GPU hosts the open-weight model locally, which is queried through a local API endpoint, whereas the closed model is accessed using Azure’s API services. In both cases, the temperature is fixed at zero. The models are probed in zero-shot setting and the maximum number of generated tokens is capped at $300$ for instruction generation and code generation.

\subsection{Task Variations}
\label{subsec:expsetup-taskvariations}
\subsubsection{Interaction Settings} We evaluate target-structure reconstruction in \textbf{single-agent} and \textbf{two-agent} settings. In the single-agent setting, a VLM receives the target structure and directly generates executable actions to reconstruct it. This setting resembles cases where a robot is given a graphical representation of the intended state and must translate it into actions without natural-language instruction exchange. It evaluates whether VLMs can interpret the target structure, reason about object placements, and complete the task without collaborative dialogue.

In the \textbf{two-agent} setting, the task is divided between two players. As described in Section~\ref{sec:methodology}, the Programmer acts as the instruction giver, and the Robot acts as the instruction follower. This setting resembles cases where a user, or another VLM-powered system, communicates a goal to a robot through instructions. It allows us to analyze how role decomposition affects task performance, target-structure interpretation and instruction quality for the Programmer, and instruction interpretation and action generation for the Robot.

We evaluate both settings under \textbf{single-turn} and \textbf{multi-turn} variants. In the single-turn variant, the agent(s) generate all actions or instructions in one response. In the multi-turn variant, the interaction proceeds iteratively: agents can use updated grid states, ask clarification questions, and generate correction instructions to refine reconstruction.

\begin{figure}[t]
  \includegraphics[width=0.49\textwidth]{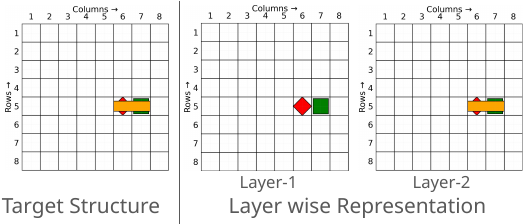}
  \caption{Example of the layer-wise target representation. The full target structure is shown alongside images that expose the structure at each stacking level.}
  \label{fig:layerwiserep}
  \vspace{-0.3cm}  
\end{figure}

\subsubsection{Target Image Representations} 
To examine how the Programmer perceives the target structure, we evaluate two image-based settings. The first uses a \textbf{single target image} showing the structure from a top-down view (Figure~\ref{fig:layerwiserep}). This setting is similar to surface mount technology (SMT) pick-and-place, where a robot observes a board from an overhead camera and infers component placements. In this view, components at intermediate stacking levels may be occluded, with only some corners or surfaces visible. This allows us to evaluate whether VLMs can accurately perceive the structure and generate corresponding instructions.

In the second setting, inspired by~\citep{DBLP:conf/iclr/SuoMSZ025, DBLP:journals/corr/abs-2505-02152, DBLP:journals/corr/abs-2505-11409} we use \textbf{layer-wise} representation for the target image. The Programmer receives the target image along with additional images showing the structure at each stacking level (see Figure~\ref{fig:layerwiserep}). For example, the first image shows placements at the ground level, the second image shows placements up to the first stacking level, and so on. These variations allow us to evaluate which visual representation is more interpretable for VLMs and how interpretation changes with structural complexity.

\begin{figure*}[t]
  \includegraphics[width=\linewidth]{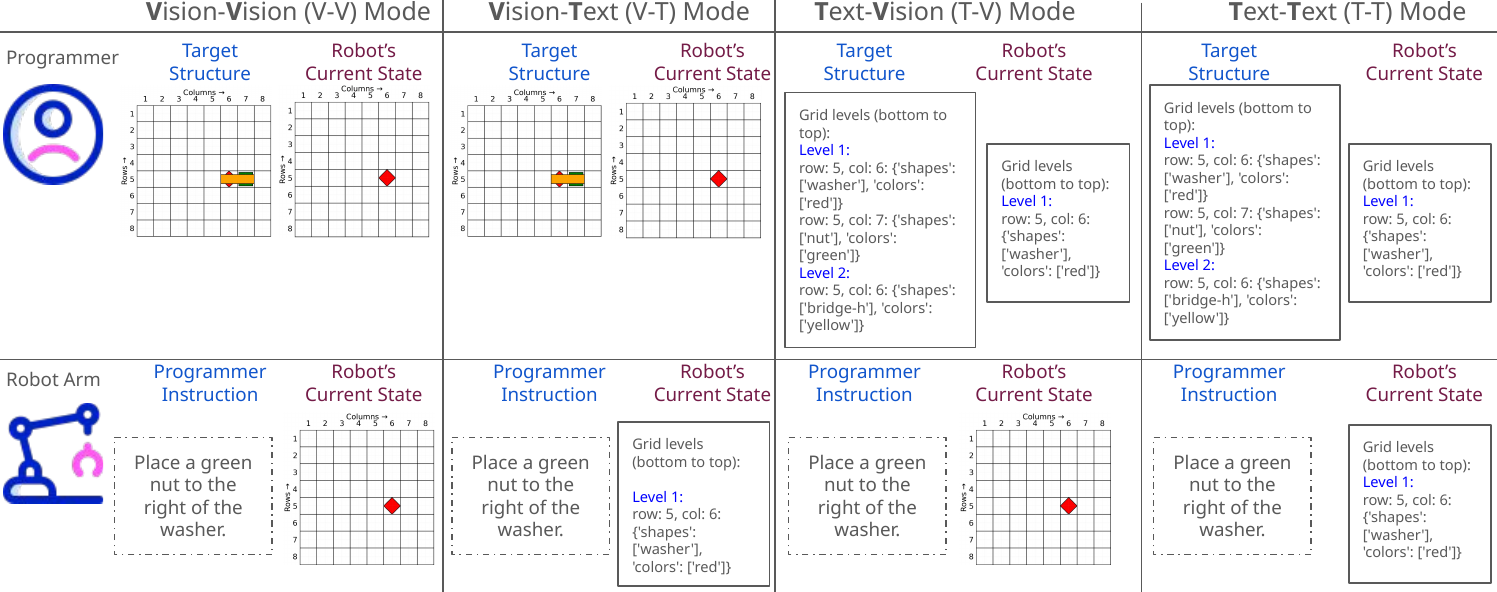}
  \caption{Example input representations for the Programmer and Robot across modality conditions. The Programmer receives the target structure and the Robot's current state, while the Robot receives the Programmer's instruction and its current state. The Programmer instruction is always provided to the Robot as text.}
  \label{fig:modalityvariation}
  \vspace{-0.3cm}
\end{figure*}

\subsubsection{Input Modalities}
To study how input modality affects interaction and task performance, we vary the information provided to each agent. The Programmer and the Robot receives input in one of two formats (see Figure~\ref{fig:modalityvariation}): \textbf{image-only} or \textbf{text-only}.

In the \textbf{text modality}, the Programmer receives textual descriptions of the target structure and the Robot's current state. These descriptions specify each component’s row, column, color, and stacking level (see T-V and T-T modes in Figure~\ref{fig:modalityvariation}). The Programmer also receives a \textbf{difference grid} (see Figure~\ref{fig:promptstructure_difference_grid} in the Appendix), which marks whether each cell in the target structure and the Robot's current state is identical, missing, or contains extra components. This modality provides explicit state information, while in the image-only modality the Programmer must infer these details from images. Therefore, the text modality serves as an upper-bound reference rather than a direct comparison to image-only input.

\subsection{Evaluation Metrics} As the main aim of the task is to reconstruct the target structure, we follow the SARTCo task evaluation protocol. At the end of the dialogue the generated structure (built by executing the code at each turn in the dialogue) is compared against the target structure in terms of each cell value (component type, color, location and stacking order). This is a binary metric referred to as \textit{Reconstruction Success Rate}, where success indicates that the generated structure exactly matches the target structure. We report the average success rate across all dialogues in the test set. A higher reconstruction success rate indicates better task performance.

\section{Results}
\subsection{Analysing the Single VLM Agent Performance}
\textit{Can a single VLM solve the structure-building task?} This question helps determine whether dialogue is necessary for the task. Table~\ref{tab:singleagent} shows the reconstruction success rate for single-turn and multi-turn interactions with a single VLM agent. Overall performance is low for both models. GPT-5.2-Chat has higher reconstruction success than Qwen3-VL-30B in both \textbf{single-turn} ($0.182$ vs. $0.036$) and \textbf{multi-turn} ($0.218$ vs. $0.036$) settings. The slight difference between single-turn and multi-turn performance suggests that the main difficulty is not the number of interaction steps, but the model's ability to interpret the target structure and generate a correct reconstruction. We therefore next examine a two-agent setting, where task roles are decomposed and agents can use clarification questions and correction instructions.

\begin{table}
  \centering
  \begin{tabular}{ccc}
    \hline
    \textbf{Model} & \textbf{Single-Turn} & \textbf{Multi-Turn}\\
    \hline
    \verb|Qwen3-VL-30B|     & 0.036 & 0.036           \\
    \verb|GPT5.2-Chat|     & 0.182 & \textbf{0.218}           \\\hline
  \end{tabular}
  \caption{Overall reconstruction success rate using a single VLM agent for the structure building task.}
  \label{tab:singleagent}
  \vspace{-0.2cm}
\end{table}

\begin{table}
  \centering
  \begin{tabular}{ccc}
    \hline
    \textbf{Model} & \textbf{Single-Turn} & \textbf{Multi-Turn}\\
    \hline
    \verb|Qwen3-VL-30B|     & 0.022 & 0.024           \\
    \verb|GPT5.2-Chat|     & 0.004 & \textbf{0.275}           \\\hline
  \end{tabular}
  \caption{Overall reconstruction success rate using two VLMs for the structure building task.
  }
  \label{tab:twoagentnolayers}
  \vspace{-0.3cm}  
\end{table}

\subsection{Analysing Two Agent Performance}
\label{subsec:results-twoagent-wolayers}
\textit{Can two VLMs collaborate to build a structure, and how does collaboration affect task performance?} We evaluate both models in self-play, where the same model is used as both the Programmer and the Robot. Table~\ref{tab:twoagentnolayers} showcases collaborative performance. Both models have low success rates in the single-turn setting, suggesting that one-shot instruction generation is not sufficient for this task; we analyze the failure patterns in Section~\ref{subsec:qualitativeanalysis}. GPT-5.2-Chat improves from $0.218$ in single-agent multi-turn to $0.275$ in two-agent multi-turn, indicating that role decomposition contributes to reconstruction. However, performance remains constrained because the Programmer observes only a top-down target image, where intermediate stacking levels may be occluded, limiting instruction quality. Qwen3-VL-30B performs poorly in both settings, with failures arising from incorrect instruction generation and spatial grounding. This suggests that role decomposition alone is insufficient when both agents struggle with target interpretation and spatial grounding. We therefore examine whether decomposed target images improve reconstruction performance.

\subsection{Analysing Image Representation Impact}
\textit{Does providing decomposed target images improve structure reconstruction?} We evaluate self-play and cross-model configurations to study the effect of the Programmer model.

\begin{table}
  \centering
  \begin{tabular}{ccc}
    \hline
    \textbf{Model} & \textbf{Single-Turn} & \textbf{Multi-Turn}\\
    \hline
    \verb|Qwen3-VL-30B|     & 0.065 & 0.180           \\
    \verb|GPT5.2-Chat|     & 0.111 & \textbf{0.493}           \\\hline
    \verb|GPT-Qwen3|     & 0.115 & 0.364           \\
    \verb|Qwen3-GPT|     & 0.091 & 0.125           \\\hline
  \end{tabular}
  \caption{Overall reconstruction success rate using two VLMs for the structure building task by using decomposed layer representation; Qwen3: Qwen3-VL-30B; GPT: GPT5.2-Chat; Qwen3-GPT/GPT-Qwen3: Indicates the first model is used as Programmer and the second model is used as Robot.
  }
  \label{tab:twoagentwithlayers}
  \vspace{-0.3cm}  
\end{table}

Table~\ref{tab:twoagentwithlayers} demonstrates reconstruction success rates with decomposed target images. Single-turn success rates improve over the single-target-image setting, but remain low, suggesting that one-shot instruction generation still limits reconstruction. In self-play multi-turn, GPT-5.2-Chat improves from $0.287$ to $0.493$, while Qwen3-VL-30B also improves to $0.180$ but remains low. In cross-model multi-turn, Qwen3-VL-30B as the Robot reaches $0.364$ with GPT-5.2-Chat as the Programmer, compared to $0.180$ in self-play. 
Conversely, GPT-5.2-Chat as the Robot drops to $0.125$ from $0.493$ in self-play when Qwen3-VL-30B acts as the Programmer. These results suggest that decomposed images improve instruction quality, and that the Programmer role affects reconstruction.

These results indicate that decomposed images help the Programmer interpret stacking relations that may be occluded in a top-down target image. The cross-model results further show that instruction generation affects collaborative reconstruction.

\subsection{Analysing Modality Impact}
\textit{How does input modality affect reconstruction?} We evaluate text-only, image-only, and text-and-image representations to examine how input format affects task performance.

\begin{table}
  \centering
  \small
  \begin{tabular}{cccc}
    \hline
    \textbf{Mode} & \textbf{Model} & \textbf{Single-Turn} & \textbf{Multi-Turn}\\
    \hline
    \multirow{2}{*}{T-V} & \verb|Qwen3-VL-30B|  & 0.067 & 0.227  \\
                                & \verb|GPT5.2-Chat|    & 0.879 & \textbf{0.910}  \\\hline
    \multirow{2}{*}{V-T} & \verb|Qwen3-VL-30B|  & 0.091 & 0.165  \\
                                & \verb|GPT5.2-Chat|    & 0.145  & \textbf{0.493}  \\\hline
    \multirow{2}{*}{T-T} & \verb|Qwen3-VL-30B|   & 0.067 & 0.364  \\
                                & \verb|GPT5.2-Chat|    & 0.911  & \textbf{0.917}  \\\hline
  \end{tabular}
  \caption{Overall reconstruction success rate using two VLMs for the structure building task with change in input modalities; T-V: Text-Vision mode; V-T: Vision-Text mode; T-T: Text-Text mode.
  }
  \label{tab:twoagentmodality}
  \vspace{-0.3cm}  
\end{table}

Table~\ref{tab:twoagentmodality} shows reconstruction success rates across modality conditions. Text-based settings lead to higher success rates than image-only settings, since they remove the need to infer structure from images. For GPT-5.2-Chat, the multi-turn Text-Image setting reaches $0.910$, and Text-Text reaches $0.917$. This suggests that when the Programmer receives the target structure in text form, the Robot can translate the resulting instructions into actions using either image-based or text-based state representations. In contrast, Image-Text reaches $0.493$, showing that the Programmer's input modality has a larger effect on reconstruction.

For Qwen3-VL-30B, the same pattern appears at lower scores. The Text-Text setting reaches $0.364$ in multi-turn, compared with $0.227$ for Text-Image and $0.165$ for Image-Text. Across both models and modality conditions, multi-turn interaction improves over single-turn, suggesting that iterative instruction exchange helps reconstruction when the input representation eases interpretation process.

\subsection{Analysing the Effect of Structural Complexity}

\textit{How does reconstruction performance change as target structures become more complex?} We analyze reconstruction success by the number of components in the target structure. Since the test set contains structures with two to five components, this analysis allows us to examine how performance changes as the number of required placements and stacking relations increases.

\begin{figure}[t]
  \includegraphics[width=0.49\textwidth]{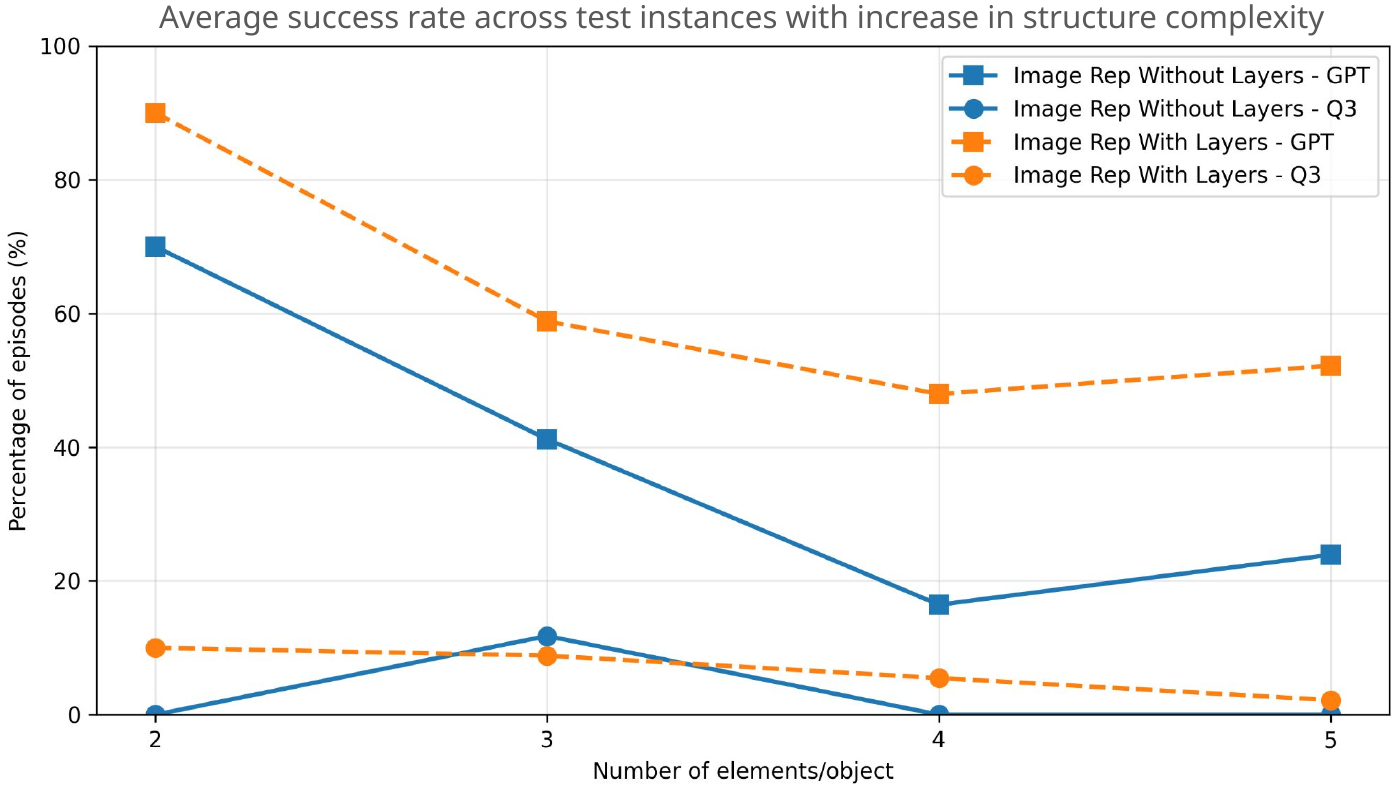}
  \caption{Reconstruction success rate across structure complexity levels in two-agent multi-turn image-based settings. Results are shown for target images with and without layer-wise representations.}
  \label{fig:overallsuccessrate}
  \vspace{-0.3cm}  
\end{figure}

Figure~\ref{fig:overallsuccessrate} shows reconstruction success across complexity levels for two-agent multi-turn image-based settings, with and without decomposed target images. For GPT-5.2-Chat, reconstruction success decreases as the number of components increases in both image representation settings, reflecting the need to reason over more component placements, stacking relations, and dependencies. Decomposed target images yield higher reconstruction success across all complexity levels, indicating that access to intermediate stacking levels helps the Programmer generate instructions for structures with more components. For Qwen3-VL-30B, reconstruction success remains low across complexity levels, suggesting difficulties with spatial interpretation and instruction following.

These results show that structural complexity affects reconstruction performance in model-dependent ways. GPT-5.2-Chat degrades with complexity but benefits from decomposed target images, while Qwen3-VL-30B remains limited across complexity levels and image representations.

\subsection{Analysing Dialogue Behaviours}
We next analyze dialogue behavior in the two-agent multi-turn image-based settings. We focus on clarification questions to understand how agents use dialogue to resolve uncertainty and recover from errors. The analysis on correction instructions is available in the Appendix~\ref{subsubsec:appendix-dlg-correction}

\begin{figure}[t]
  \includegraphics[width=0.48\textwidth]{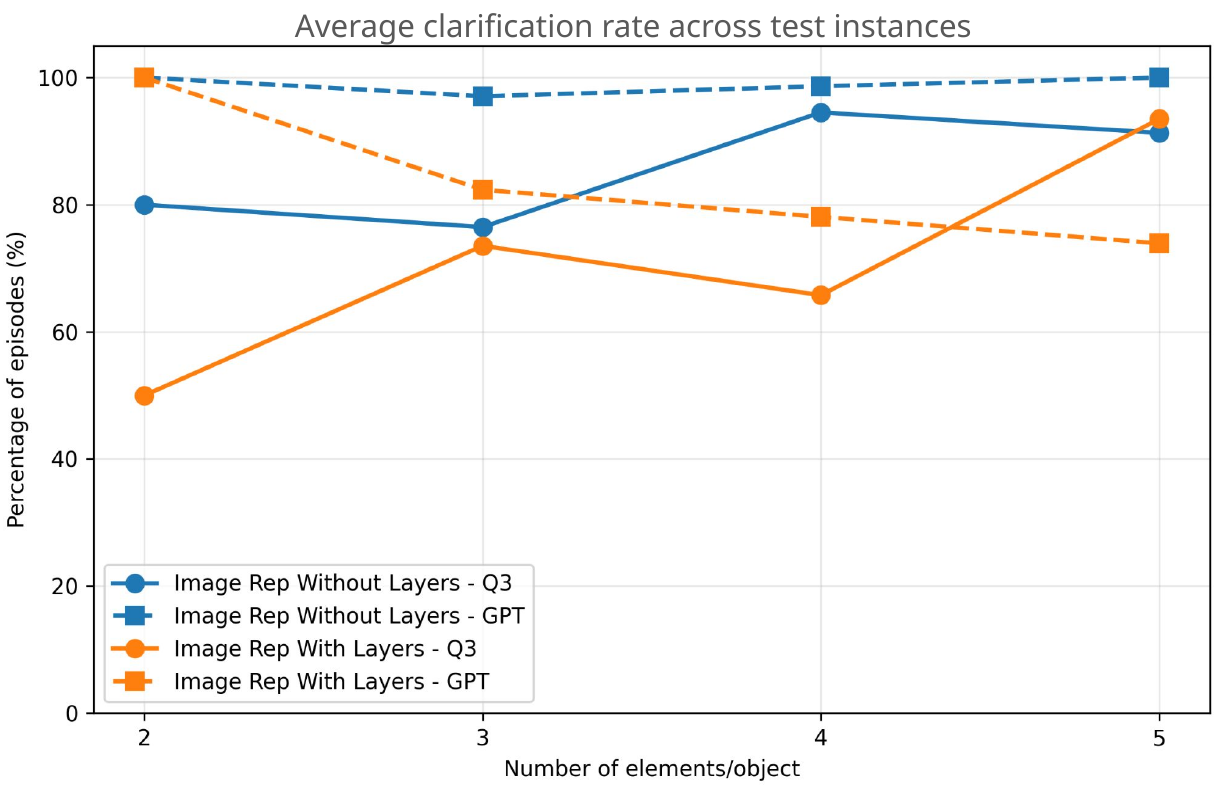}
  \caption{Percentage of two-agent multi-turn episodes containing clarification questions across structure complexity levels, with and without layer-wise target representations.}
  \label{fig:cfqcounts}
  \vspace{-0.3cm}  
\end{figure}

\begin{figure}[t]
  \includegraphics[width=0.49\textwidth]{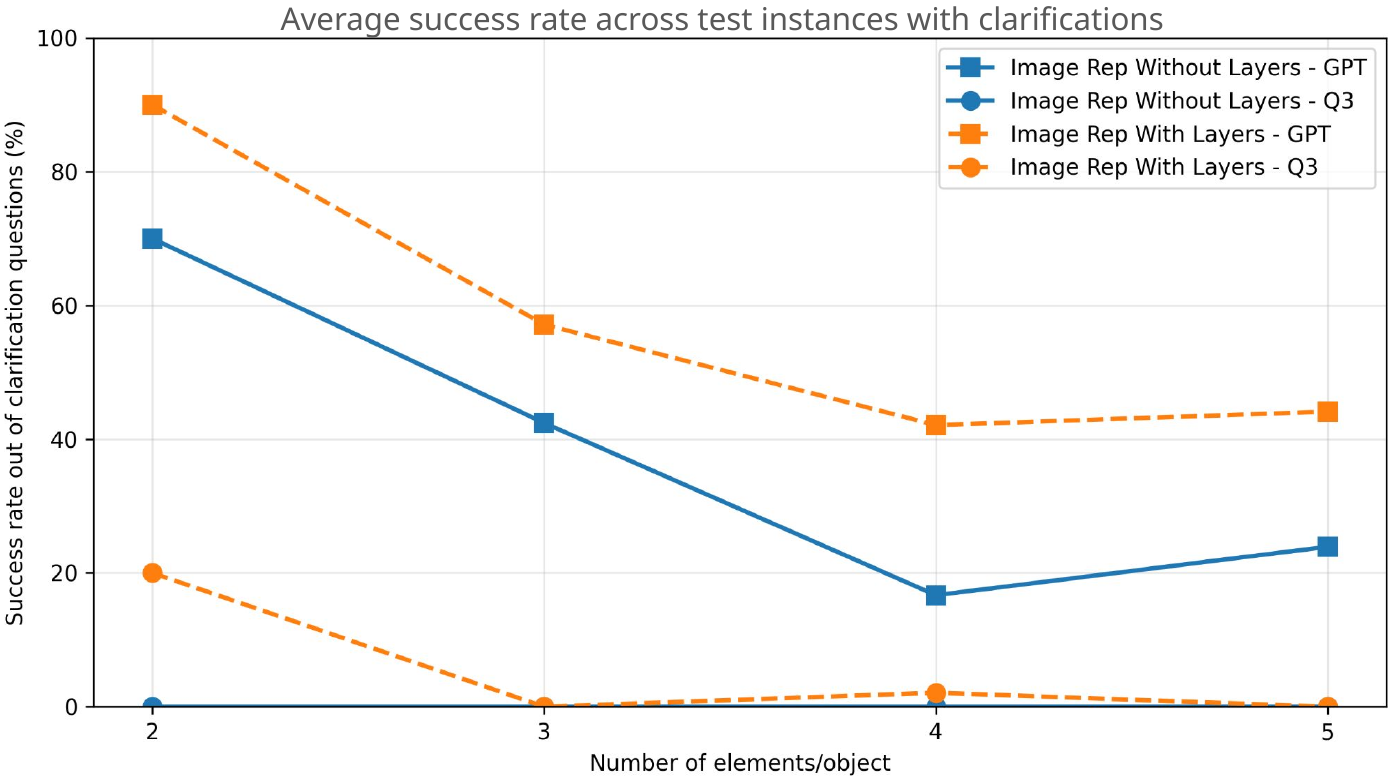}
  \caption{Reconstruction success rate for two-agent multi-turn episodes that contain clarification questions, evaluated across structure complexity levels with and without layer-wise target representations.}
  \label{fig:cfqsuccess}
  \vspace{-0.3cm}
\end{figure}

Clarification questions indicate cases where the Robot cannot translate the Programmer's instruction into executable actions, either because the instruction is underspecified, or violates environment constraints. Figure~\ref{fig:cfqcounts} shows the percentage of episodes that contain at least one clarification question across structure complexity levels.

Without decomposed target images, clarification questions occur in most episodes for both models, with rates close to $97$--$100\%$ across complexity levels. This suggests that top-down target images often lead to instructions that require additional grounding before execution. With decomposed target images, the pattern changes. For GPT-5.2-Chat, the clarification rate decreases as structure complexity increases, while for Qwen3-VL-30B, the clarification rate increases with complexity. This difference suggests that the models use clarification questions differently.

\begin{figure*}[t]
  \includegraphics[width=\textwidth]{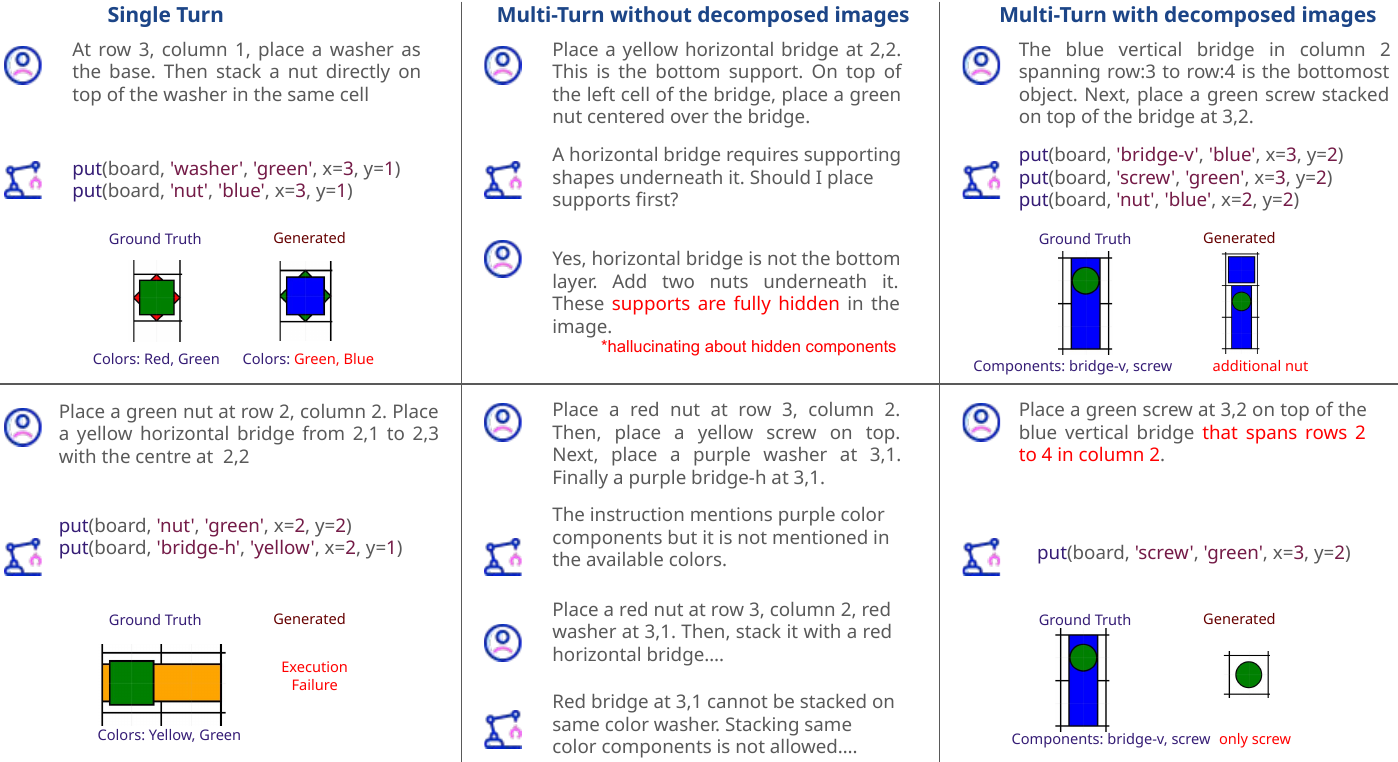}
  \caption{Two-agent failure cases for GPT-5.2-Chat (top-panel) and Qwen3-VL-30B (bottom-panel) across interaction settings. Failures include missing color information, depth mismatches, clarification loops, and incorrect reconstruction from misinterpreted target structures.}
  \label{fig:qualitative-analysis}
  \vspace{-0.3cm}
\end{figure*}

We now analyse how clarification questions relate to task success. Figure~\ref{fig:cfqsuccess} shows reconstruction success among episodes that contain clarification questions. Qwen3-VL-30B has near-zero success across image representations, indicating that clarification does not lead to task recovery for this model. GPT-5.2-Chat achieves higher success when decomposed target images are used, especially for lower-complexity structures. However, success decreases as the number of components increases, suggesting that clarification alone does not resolve failures caused by visual misinterpretation or incorrect spatial grounding. Overall, decomposed target images contribute more to successful recovery than clarification behavior alone.

\subsection{Qualitative Analysis}
\label{subsec:qualitativeanalysis}
We analyze two-agent interactions to identify failure patterns (see Figure~\ref{fig:qualitative-analysis}). In GPT-5.2-Chat single-turn interactions, $93\%$ of the executed interactions contain color mismatches between the generated code and the ground truth. This occurs because the Programmer omits color information in the instructions, forcing the Robot to assume colors. For Qwen3-VL, $75.6\%$ of episodes abort due to incorrect stacking depths, missing color information, and component-specific constraint violations.

In the GPT-5.2-Chat multi-turn setting with a single target image, some instructions describe structures that differ from the target. Robot clarification questions often surface these errors, but the exchange does not always recover as Programmer's target interpretation is incorrect. For Qwen3-VL, multi-turn interaction reduces aborts, with only seven test episode resulting in an abort. However, $79.6\%$ of interactions reach the $15$-turn limit because the Robot repeatedly asks clarification questions without generating executable actions. This suggests that multi-turn interaction can reduce execution aborts but may lead to clarification loops.

Even with decomposed images, GPT-5.2-Chat still misinterprets some target structures, leading to execution failures in $20\%$ of instances. Even when the target is interpreted correctly, the Robot misinterprets instructions, causing incorrect stacking order, extra components or location errors. For Qwen3-VL, $54\%$ of interactions reach the $15$-turn limit. The Programmer sometimes treats decomposed images as separate targets and repeats construction instructions, while the Robot fails to preserve stacking order or object locations.

\section{Conclusion}
This paper explores the collaborative capabilities of VLMs in a structure-building task. To this end, we evaluated single-agent and two-agent settings across multiple modalities and image representations. Our findings show that spatial reasoning remains challenging for the evaluated VLMs. Two-agent collaboration alone provides limited gains, suggesting that role decomposition does not resolve errors in visual interpretation and spatial grounding. We observe larger improvements in task performance when intermediate stacking-layer image representations are provided. Furthermore, text-based target representations lead to higher reconstruction success across all evaluated modalities. We also find that clarification and correction behaviors do not always lead to recovery, indicating that failures are often caused by incorrect visual grounding rather than dialogue alone. Future work should focus on improving visual spatial grounding and grounded instruction generation for VLM agents in collaborative structure-building tasks.

\section*{Limitations}
Although the proposed framework enables a controlled study, it has several limitations. First, our experiments focus on simple boards that have a single object on the grid. Although this setup allows us to isolate failures at different stages of the interaction, it does not cover the full range of perception and manipulation challenges in physical robot environments. Second, we use 2.5D target representations, which do not capture the properties of full 3D representations or their effects on model performance. Third, this work is an illustrative case study rather than a comprehensive benchmark, and evaluates two representative VLMs across open-weight and closed paradigms. Broader generalization of our findings to VLMs as a class would require evaluation across a wider range of models. Future work can extend the evaluation to physical robot settings, broader model families, and 3D representations.


\bibliography{custom}

\appendix

\section{Appendix}
\label{sec:appendix}

\subsection {Interaction Framework}
\label{subsec:appendix-interaction-fw}
As described in Section~\ref{subsec:methodology-vlmselfplay}, we build the two-player interaction using the \texttt{clem:todd} framework~\citep{kranti-etal-2025-clem}. Player~1, the \textit{Programmer}, acts as the instruction giver, similar to a simulated user. Player~2, the \textit{Robot}, acts as the instruction follower and translator, similar to a dialogue system. Both players are VLMs. In self-play, the same VLM plays both roles, while in cross-play, different VLMs are assigned to the Programmer and Robot roles.

The interaction is orchestrated by a Game Master (GM), which handles turn-level coordination. The GM first prompts the Programmer, which is expected to return only a natural-language instruction marked by response anchors; details on the anchors are available in Appendix~\ref{subsubsec:appendix-prompt-templates}. The GM removes the anchors, verifies that the response is textual, and passes the instruction to the Robot. The Robot is prompted to return a JSON response with two keys: \texttt{status} and \texttt{details}. The \texttt{status} field can take one of three values: \texttt{code}, \texttt{clarification}, or \texttt{acknowledgement}. The \texttt{details} field contains the corresponding code, clarification question, or acknowledgement text. The GM removes anchors, validates the JSON structure, checks whether the \texttt{status} value is valid, extracts the \texttt{details} field, and routes the response to the next step.

If the Robot returns \texttt{code}, the GM sends the code to a virtual simulator, which executes it and produces an updated grid-state image. The GM then passes this image to the Programmer as the result of the Robot's action. If execution fails, the GM sends the error feedback to the Robot so that it can revise the code. If the Robot returns \texttt{clarification} or \texttt{acknowledgement}, the GM forwards the response to the Programmer without modification. The Programmer does not observe the Robot-generated code directly; it only observes the updated grid state after execution.

\begin{figure*}[t]
  \includegraphics[width=\linewidth]{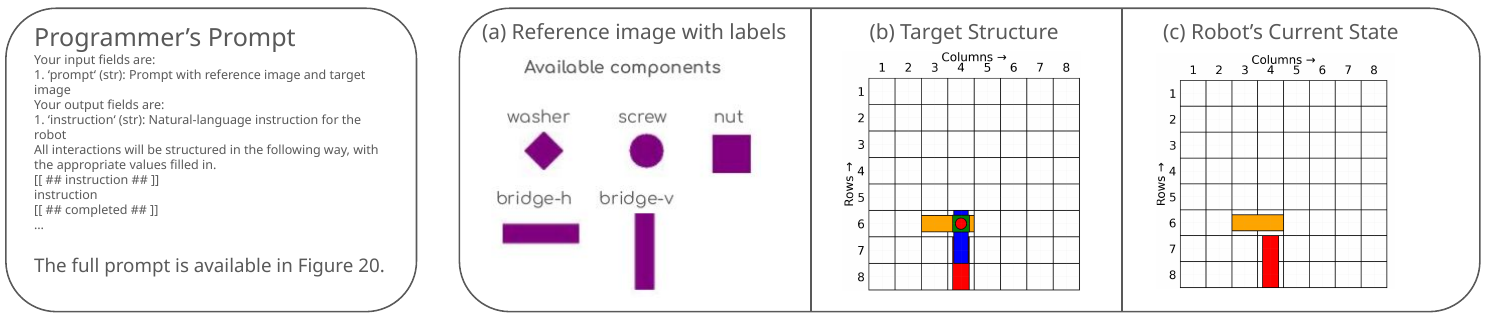}
  \caption{Programmer Prompt Overview}
  \label{fig:programmerpromptoverview}
  \vspace{-0.3cm}
\end{figure*}

\begin{figure*}[t]
  \includegraphics[width=\linewidth]{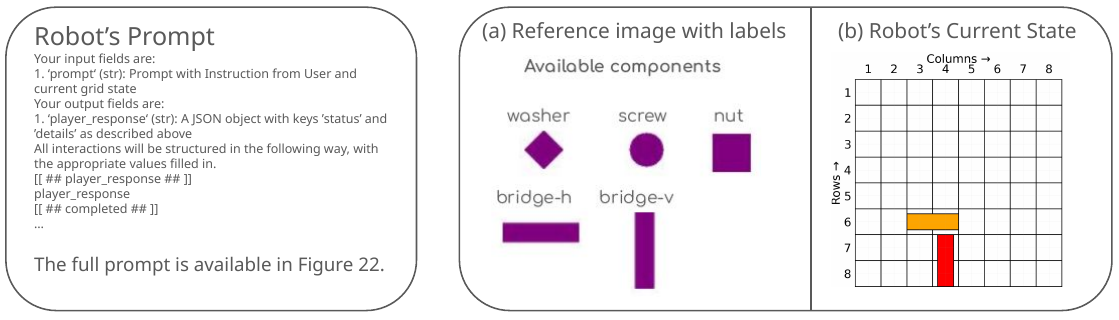}
  \caption{Robot Prompt Overview}
  \label{fig:robotpromptoverview}
  \vspace{-0.3cm}
\end{figure*}

The interaction ends when the Programmer outputs \texttt{DONE} or when the turn limit is reached. After termination, the GM compares the Robot's final grid state with the target structure and marks the episode as successful only if they match exactly. The comparison is performed at the cell level and checks component type, color, location, and stacking order.

\paragraph{Virtual Simulator} The virtual simulator executes the Robot-generated Python code and produces both an image and an ASCII representation of the Robot's current grid state. Depending on the experimental setting, the GM passes the relevant representation to the Programmer. The simulator also provides APIs to compute an ASCII difference grid between the target structure and the Robot's current grid state, marking matches, missing components, and extra components at each stacking level. We use this difference grid in the Text-Text setting to provide explicit feedback about the Robot's current grid state to the Programmer.

\subsubsection {Prompt Templates}
\label{subsubsec:appendix-prompt-templates}
Following standard prompting approaches~\citep{DBLP:conf/nips/BrownMRSKDNSSAA20, DBLP:conf/nips/Wei0SBIXCLZ22, DBLP:journals/csur/LiuYFJHN23}, we use zero-shot prompts for all experiments. The prompts vary by agent role and experimental setting, but follow a common structure: system message, task description, environment information, rules, constraints, and the previous response from the other player. Inspired by DSPy~\citep{DBLP:journals/corr/abs-2310-03714}, we add response anchors \texttt{[[ \#\# instruction \#\# ]]}, \texttt{[[ \#\# player\_response \#\# ]]} and \texttt{[[ \#\# completed \#\# ]]} to have consistent model outputs.

The Programmer prompt describes the task, environment information, and constraints. It also includes a reference image with component labels, the target structure, and the Robot's current state (see Figure~\ref{fig:programmerpromptoverview}). The Robot prompt contains the task description, environment information, constraints, the Programmer's instruction, a reference image with component labels, and the Robot's current state (see Figure~\ref{fig:robotpromptoverview}). The response format differs by role: the Programmer produces a natural-language instruction, while the Robot returns a JSON response containing \texttt{status} and \texttt{details}.

\begin{table*}
\centering
\begin{tabular}{ccccccc}
\hline
\multirow{2}{*}{Model} & \multicolumn{3}{c}{Single Turn} & \multicolumn{3}{c}{Multi-Turn} \\
                       & Abort   & Failure   & Success   & Abort   & Failure   & Success  \\ \hline
\verb|Qwen3-VL-30B| & 0.604 & 0.360 & 0.036 & 0.913 & 0.051 & 0.036  \\
\verb|GPT5.2-Chat|  & 0.374 & 0.444 & \textbf{0.182} & 0.479 & 0.303 & \textbf{0.218}\\ \hline
\end{tabular}%
\caption{Detailed statistics for single agent settings}
\label{tab:appendix-sagent-detailedstats}
\end{table*}

\paragraph{Ablation Studies} We conducted preliminary experiments to refine the prompt templates and response formats. Based on these observations, we added \texttt{clarification} and \texttt{acknowledgement} as Robot response statuses, allowing the Robot to indicate ambiguity or acknowledge an instruction in addition to code generation. We also added execution-error feedback for the Robot, allowing up to three revision attempts before aborting an episode. This feedback allows the Robot to correct code errors or ask relevant clarification questions. We fixed the maximum tokens at $300$ tokens and interaction turns as $15$. In preliminary runs, this limit did not lead to response truncation, dialogue aborts, or incorrect outputs, so we used it across all final experiments.

\begin{table*}
\centering
\begin{tabular}{ccccccc}
\hline
\multirow{2}{*}{Model} & \multicolumn{3}{c}{Single Turn} & \multicolumn{3}{c}{Multi-Turn} \\
                       & Abort   & Failure   & Success   & Abort   & Failure   & Success  \\ \hline
\verb|Qwen3-VL-30B|& 0.756 & 0.222 & \textbf{0.022} & 0.913 & 0.063 & 0.024 \\
\verb|GPT5.2-Chat|& 0.372 & 0.624 & 0.004 & 0.386 & 0.339 & \textbf{0.275} \\ \hline
\end{tabular}%
\caption{Detailed statistics for two VLM agents (V-V mode with single target image) }
\label{tab:appendix-twoagent-detailedstats}
\end{table*}

\subsection {Dataset}
\label{subsec:appendix-dataset}
As described in Section~\ref{sec:expsetup}, we use goal structures from the SARTCo dataset~\citep{kranti2024towards}. The dataset represents boards as arrangements of components on a 2.5D $8 \times 8$ grid. The components mimic assembly settings and include washers, nuts, horizontal bridges, vertical bridges, and screws. Each component is available in four colors: red, blue, green, and yellow. Components can be connected on the grid in arrangements that conceptually allow electricity to flow through them. Each such connected arrangement is referred to as an \textit{object}, and the placement of objects on the grid is referred to as a \textit{board}. Each object contains between two and five components.

The component arrangements are represented using Python programs. Each board is also paired with synthetic instructions generated from a template grammar. Thus, each board is associated with both Python code and synthetic instructions that describe the same target structure.

The component arrangements follow predefined constraints: (a) horizontal and vertical bridges occupy two cells, while other components occupy one cell; (b) cells support vertical stacking; (c) a screw can only be placed at the top, and no other component can be stacked on it; (d) bridge placement must satisfy depth constraints; and (e) components with the same color or type cannot be stacked on each other.

The boards are categorized into two types based on the number of objects on the board. Boards with a single object are grouped as simple boards, while boards with multiple objects arranged in different patterns are grouped as regular boards.

In this work, we use only the target structures from the simple boards. The simple-board subset is split into train, validation, and test sets. The train set contains 165 boards, while the validation and test sets contain 495 boards each. We use the test set for all experiments. We execute the Python code associated with each board to render the component arrangement using Matplotlib, and use the generated images as goal structures for our task. We also use the Python code during evaluation to compare cell values between the target structure and the Robot-generated structure, including component type, color, location, and stacking order. Although the dataset includes synthetic instructions, we did not use them because instructions are generated by the instruction giver (Programmer) during interaction.

\subsection {Quantitative Analysis}
\label{subsec:appendix-quant-analysis}
In addition to reconstruction success, we analyze interaction aborts and execution failures to better understand the interaction. An episode is marked as aborted if either player's response violates the required response constraints, or if the Robot-generated code produces execution errors for three consecutive attempts. All episodes that are not aborted are compared against the target structure. An episode is marked as successful only if the final structure matches the target exactly; otherwise, it is marked as a failure.

\begin{table*}
\centering
\begin{tabular}{ccccccc}
\hline
\multirow{2}{*}{Model} & \multicolumn{3}{c}{Single Turn} & \multicolumn{3}{c}{Multi-Turn} \\
                       & Abort   & Failure   & Success   & Abort   & Failure   & Success  \\ \hline
\verb|Qwen3-VL-30B|& 0.691 & 0.244 & 0.065 &  0.014 & 0.806 & 0.180 \\
\verb|GPT5.2-Chat|& 0.236 & 0.653 & 0.111 & 0.188 & 0.319 & 0.493 \\ \hline
\end{tabular}%
\caption{Two VLM Agents (V-V mode with layerwise representation) Detailed Statistics}
\label{tab:appendix-twoagent-withlayers-detailedstats}
\end{table*}

\subsubsection {Single Agent Setting}
\label{subsubsec:appendix-quant-sa-st-detailstats}
\paragraph{Single-Turn} In the single-agent, single-turn setting, $60\%$ of Qwen3-VL-30B episodes are aborted. Among these aborts, $54\%$ are caused by depth mismatches during bridge placement, and $83\%$ of these occur for structures with four or five components. Among the failed episodes, $91.6\%$ contain location mismatches between the generated and target structures.

For GPT-5.2-Chat, $37\%$ of episodes are aborted. The main causes are invalid bridge placement at the top level ($39\%$), horizontal-bridge depth mismatches ($22.7\%$), and same-color stacking errors ($21.6\%$). Among failed episodes, $66\%$ contain location mismatches, while $15.5\%$ contain extra components in the generated structure. Other failures include missing components, color mismatches, and stacking-order errors.

Among successful Qwen3-VL-30B episodes, $67\%$ are from two-component structures and the rest are from three-component structures. For GPT-5.2-Chat for two-components $56.7\%$ of the instances are successful, for three components $31.4\%$  of the instances are successful, for four components $13.2\%$  of the instances are successful, and for five components $8.3\%$ of the instances are successful.

\paragraph{Multi-Turn} In the single-agent, multi-turn setting, $91.3\%$ of Qwen3-VL-30B episodes are aborted. Among these aborts, $37.1\%$ are caused by not following environment constraints (stacking same shapes, on top of screw) and $35\%$ are caused by depth mismatches during bridge placement.

For GPT-5.2-Chat, $47.9\%$ of episodes are aborted. The main causes are due to not following environment constraints ($51.9\%$) followed by $17.7\%$ depth-mismatch errors for horizontal bridges. Among failed episodes, $63.1\%$ contain location mismatches, while $14.8\%$ contain extra components in the generated structure. Other failures include missing components, color mismatches, and stacking-order errors.

Among successful Qwen3-VL-30B episodes, $55.5\%$ are from two-component structures and the rest are from three-component structures. For GPT-5.2-Chat, for two-components $83.33\%$ are successful, for three components $40.19\%$ are successful,  four components $14.61\%$ are successful, and for five components $6.9\%$ structures are successful. 

Overall, in the single-agent settings, both models struggle with element count, location, color, and stacking order, especially for structures with three or more components. This suggests that top-down target images are difficult to interpret in one turn, where the model cannot revise its reconstruction based on feedback.

\subsubsection {Two Agent Settings}
\label{subsubsec:appendix-quant-twoa-detailstats}
\paragraph{Single-Turn} We now analyse the two-agent interaction with single target image representation. In the single-turn setting, $75.6\%$ of Qwen3-VL-30B episodes are aborted. Among these aborts, $65.7\%$ are caused by depth mismatches during bridge placement, and $76.35\%$ of these occur for structures with four or five components. Among the failed episodes, $91.6\%$ contain location mismatches between the generated and target structures.

For GPT-5.2-Chat, $37\%$ of episodes are aborted. The main causes are same-color stacking errors ($37.5\%$) and invalid bridge placement at the top level ($35.3\%$). Among failed episodes, $93\%$ contain color mismatches, while $6.1\%$ contain extra components in the generated structure. Other failures include location mismatches, and stacking-order errors.

\paragraph{Multi-Turn} In the multi-turn setting, $91.3\%$ of Qwen3-VL-30B episodes are aborted. Among these aborts, $37.2\%$ are caused by depth mismatches during bridge placement and $40.0\%$ are caused by not following environment constraints (stacking same shapes, stacking on top of screw, placing bridges at the top-most layer).

For GPT-5.2-Chat, $38.6\%$ of episodes are aborted. Out of these aborts, $37.2\%$ are caused by depth mismatches during bridge placement and $40.0\%$ are caused by not following environment constraints (stacking same shapes, stacking on top of screw, placing bridges at the top-most layer).

Among successful Qwen3-VL-30B episodes, $55.5\%$ are from two-component structures and the rest are from three-component structures. For GPT-5.2-Chat, for two-components $93.33\%$ are successful, for three components $42.16\%$ are successful,  four components $21.92\%$ are successful, and for five components $11.8\%$ structures are successful.

\begin{figure}[t]
  \includegraphics[width=0.48\textwidth]{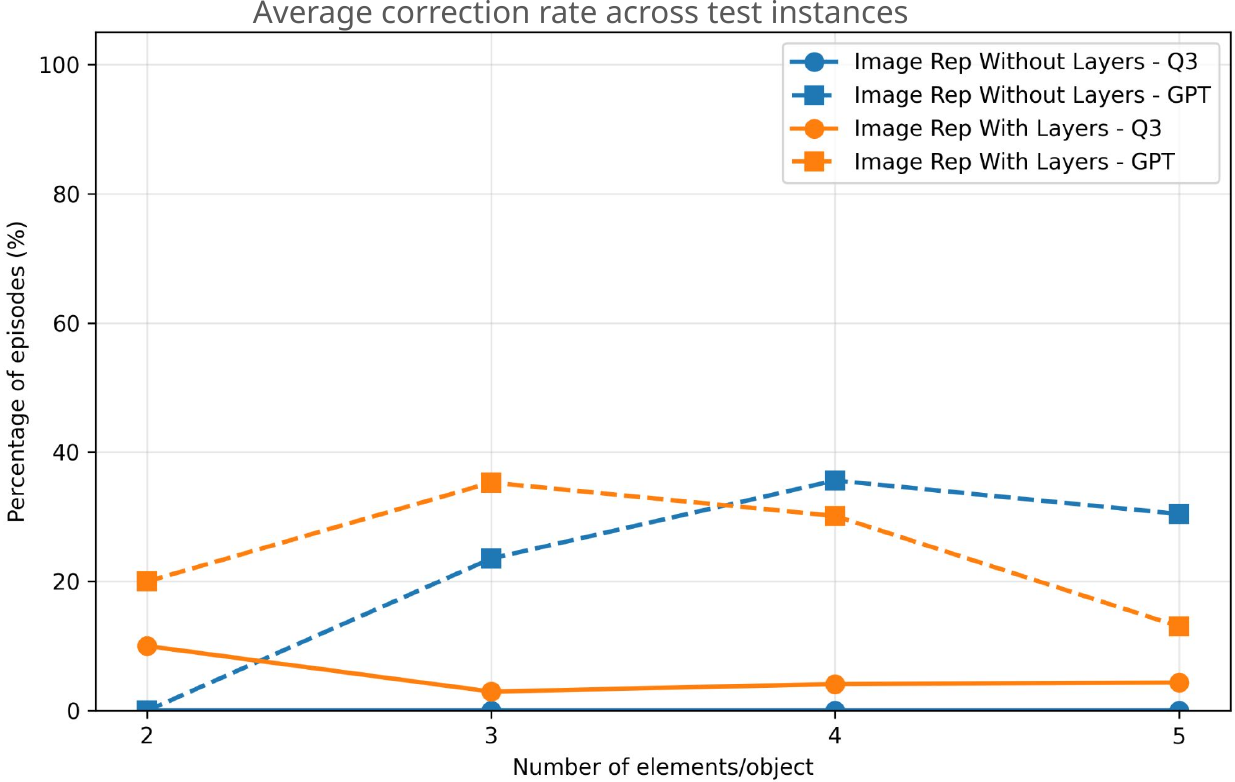}
  \caption{Percentage of two-agent multi-turn episodes containing corrections across structure complexity levels, with and without layer-wise target representations.}
  \label{fig:corrcounts}
\end{figure}

\subsubsection {Analysing Dialogue Behaviours}
\label{subsubsec:appendix-dlg-correction}
We analyze correction instructions in the two-agent multi-turn image-based settings. Correction instructions occur when the Programmer observes the updated Robot grid and attempts to revise an incorrect or incomplete action. Figure~\ref{fig:corrcounts} shows the percentage of episodes that contain correction instructions across structure complexity levels.

GPT-5.2-Chat produces correction instructions more often than Qwen3-VL-30B. Qwen3-VL-30B produces no correction instructions in either image representation, suggesting that it does not identify or act on errors in the updated grid state. For GPT-5.2-Chat, correction rates vary with image representation and structure complexity. For two- and three-component structures, correction instructions occur more often with decomposed target images than with top-down images. For four- and five-component structures, correction instructions occur more often without decomposed target images.

\begin{figure}[t]
  \includegraphics[width=0.48\textwidth]{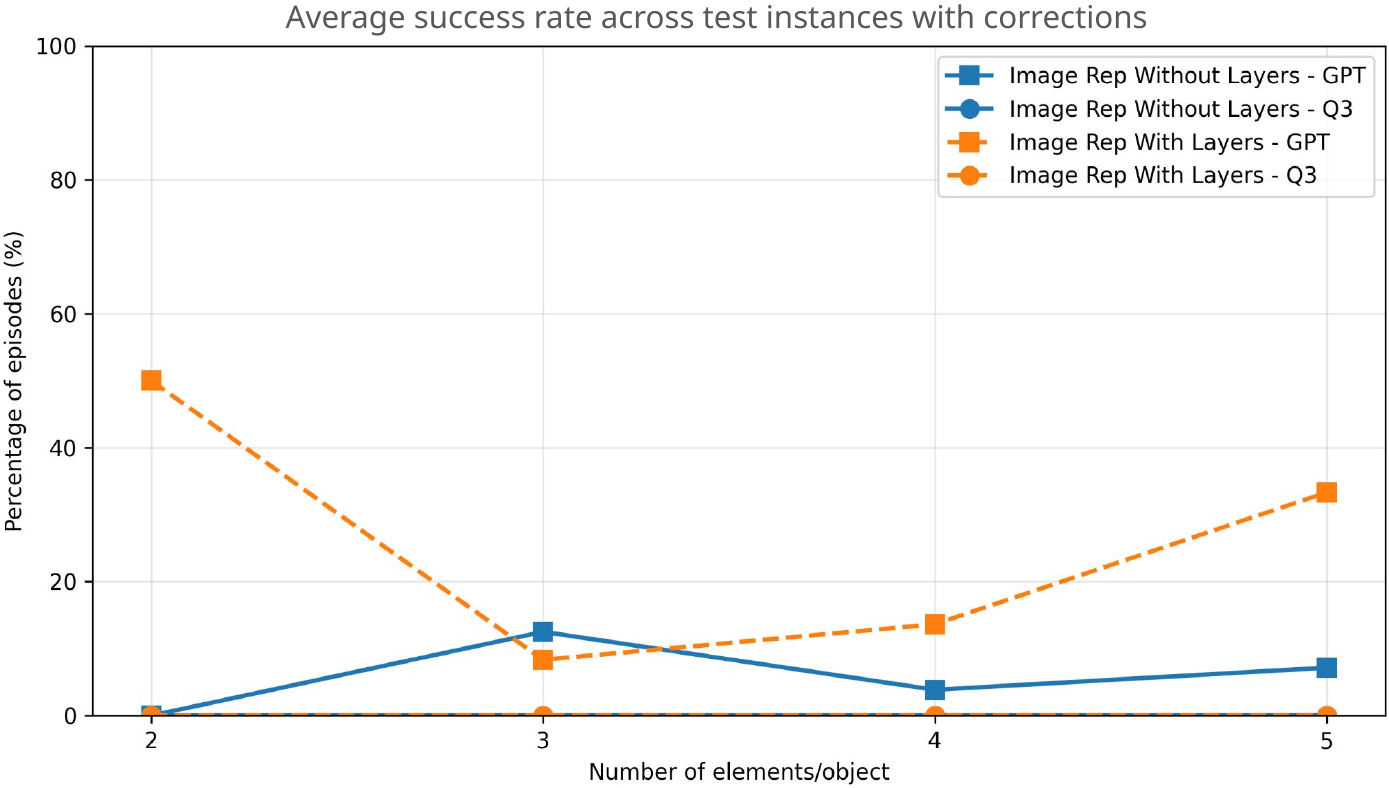}
  \caption{Reconstruction success rate for two-agent multi-turn episodes that contain corrections, evaluated across structure complexity levels with and without layer-wise target representations.}
  \label{fig:corrsuccess}
\end{figure}

Figure~\ref{fig:corrsuccess} reports reconstruction success among episodes with correction instructions. Qwen3-VL-30B has no correction episodes, so no recovery through correction is observed for this model. GPT-5.2-Chat achieves higher reconstruction success in correction episodes when decomposed target images are used. This suggests that corrections are more effective when the Programmer has access to visual representations that expose intermediate stacking levels. However, success remains limited for structures with more components, indicating that correction alone does not resolve failures caused by earlier visual or spatial reasoning errors.

\begin{figure*}
  \centering 
    \begin{prompt}
\\
Your input fields are:

1. `prompt` (str): Prompt with goal grid

Your output fields are:

1. `player\_response` (str): A JSON object with keys 'status' and 'details' as described above

All interactions will be structured in the following way, with the appropriate values filled in.

[[ \#\# player\_response \#\# ]]

{player\_response}

[[ \#\# completed \#\# ]]
\\\\
In adhering to this structure, your objective is: 

        You are the robot in a structure reconstructing game.
\\\\
        You will be shown two images:
        
        • Image 1 is a reference legend — it shows each shape alongside its label. Use this to identify and name shapes correctly that appear in Image 2.
        
        • Image 2 is the gold truth — the target structure that must be reconstructed.
        
        • Using the labels from Image 1, interpret Image 2 accurately. Identify each placed shape, reason carefully about its grid position, occupied cells, and stacking order. Then generate action sequence to reconstruct the target structure.
        The response should be a JSON object as described in the Output Format section below.        
\\\\
        Grid Orientation
        
        • Rows are numbered 1 to 8, top to bottom
        
        • Columns are numbered 1 to 8, left to right
        
        • A cell is always described as row, column — strictly in that order.
        
            Do not confuse rows and columns.
            
            row:3, col:4 and row:4, col:3 are different cells and must not be confused. 
\\\\
        Shape identification
        
        • Nut: a square shape occupying exactly one cell.
        
        • bridge-h: a horizontal bar occupying two adjacent cells in the same row.
        
        • Do not interpret a bridge-h as two separate nuts placed side by side. If relevant, also distinguish vertical extensions similarly when visible.
\\\\
        Interpretation Mode:
        
        • Always generate actions from bottom to top: supporting objects must be placed before any objects that rest on them.

        Ordering Constraints (MANDATORY):
        
        • Actions must respect stacking dependencies.
        
        • Always place supporting (lower-level) shapes before shapes that rest on them.
        
        • Never place a top-level shape before its support exists.
\\\\
        Partially occluded stacked shapes handling:
        
        • When shapes are stacked, lower shapes may be partially or fully hidden. Use these cues to identify them.
        
        • Infer the hidden shape from visible evidence:
        
            Visible corners — If only the corners of a diamond are visible, the hidden shape is likely a washer.
            
            Half of a shape extending — If part of a shape extends into the next cell horizontally, it is likely a bridge-h.
            
            Half of a shape extending — If part of a shape extends into the next cell vertically, it is likely a bridge-v.
            
            When colors match across layers, do not assume it is a single shape — interpret the stack carefully in bottom-to-top order
\end{prompt}
\caption{Part-1 of the prompt template used for the single agent VLM in single-turn setting.}
    \label{fig:promptstructure_sagent_part1}
\end{figure*}

\begin{figure*}
  \centering
    \begin{prompt}
\\
        Available APIs:
        
        - put(board: np.ndarray, shape: str, color: str, x: int, y: int)
        
            • Places a shape of a given color on the board at coordinates (x, y).

        - move(board: np.ndarray, x1: int, y1: int, x2: int, y2:int, shapes\_list: List = None)
        
            • To move a shape (or stack of shapes) from coordinates (x1, y1) to coordinates (x2, y2) on the board
            
            • Do not use same coordinates for source and destination
            
            • Do not use this function to remove shapes from the board, use removeshape instead
            
            • Use x1, y1 as source coordinates and x2, y2 as destination coordinates
            
            • Use x1, y1, x2 and y2 do not use same names for source and destination
            
            • x1, x2: Row values
            
            • y1, y2: Column values
            
            • shape\_list: (List, optional): List of shapes from bottom → top to move. If None, only the topmost shape in the stack is moved.

        - removeshape(board: np.ndarray, x: int, y: int, shape, color)
        
            • To remove a shape from coordinates (x1, y1) on the board
            
            • Do not use this function to move shapes, use move instead
            
            • Use this function only when there is a placement error. If a shape already exists in the target cell, ask whether it should be removed before placing a new one. In most cases, you will be stacking a new shape in the same cell, which doesn’t require removeShape();just call put() for the new shape.
            
            • The shape and color must match the topmost shape in the stack at (x, y)                
            • x: Row  (If you want to remove a shape from "row = 3", then use x = 3)
            
            • y: Column  (If you want to remove a shape from "column = 1", then use y = 1)
            
            • shape\_list: (List, optional): List of shapes from bottom → top to move. If None, only the topmost shape in the stack is moved.

        - clear(board: np.ndarray)
        
            • To clear all the shapes on the board

        - undo(board: np.ndarray)
        
            • To undo last placement action(s) on the board
            
            • If multiple shapes were placed as part of the previous operation, all of them will be cleared from the board.   
\\\\
        ---
        AVAILABLE SHAPES
        
          - washer
          
          - nut
          
          - screw
          
          - bridge (vertical, horizontal)  ← special shape
          
             • Use **'bridge-h'** for horizontal bridges and **'bridge-v'** for vertical ones.      
        
        AVAILABLE COLORS
        
          - green
          
          - red
          
          - blue
          
          - yellow
\\\\
        SHAPE OCCUPANCY
        
          • All shapes occupy **exactly one cell**, **except the "bridge"**.
          
          • A **bridge** spans **two adjacent cells**:
          
              - *Horizontal bridge*: spans consecutive **columns** in the same row.
              
              - *Vertical bridge*: spans consecutive **rows** in the same column.
              
          • A bridge **requires two other shapes** underneath it (referred as depth) for support while stacking, one under each end of the bridge.

\end{prompt}
\caption{Part-2 of the prompt template used for the single agent VLM in single-turn setting.}
    \label{fig:promptstructure_sagent_part2}
\end{figure*}

\begin{figure*}
  \centering
    \begin{prompt}
\\
        BRIDGE REPRESENTATION RULES
        
          • Bridges span two cells and appear in the target grid as:
          
              - bridge-v-top / bridge-v-bottom  → one vertical bridge
              
              - bridge-h-left / bridge-h-right  → one horizontal bridge
              
              - 'bridge-h-left' and 'bridge-h-right' are display labels representing the two halves of a single horizontal bridge (bridge-h). A horizontal bridge always occupies two adjacent cells in the same row. When generating code or interpreting grid , always refer to the shape as bridge-h, not the left/right variants.
              
              - Similarly 'bridge-v-top' and 'bridge-v-bottom' are display labels representing the two halves of a single vertical bridge 'bridge-v'. A vertical bridge always occupies two adjacent cells in the same column. When generating code or interpreting grid , always refer to the shape as bridge-v, not the top/bottom variants.
              
          • Example:
          
              → Code: put(board, 'bridge-h', 'green', 8, 1)
              
              → Current-Grid would become:
              
                Grid levels (bottom to top):
                
                Level 1:
                
                row: 8, col: 1: {'shapes': ['bridge-h-left'], 'colors': ['green']}
                
                row: 8, col: 2: {'shapes': ['bridge-h-right'], 'colors': ['green']}
                
              → To remove this shape: removeshape(board, x=8, y=1, shape='bridge-h', color='green')

              → Code: put(board, 'bridge-v', 'blue', 2, 8)
              
              → Current-Grid would become:
              
                Grid levels (bottom to top):
                
                Level 1:
                
                row: 2, col: 8: {'shapes': ['bridge-v-top'], 'colors': ['blue']}
                
                row: 3, col: 8: {'shapes': ['bridge-v-bottom'], 'colors': ['blue']} 
                
              → To remove this shape: removeshape(board, x=2, y=8, shape='bridge-v', color='blue')  

        STACKING \& DEPTH RULES
        
          • Shapes can be stacked vertically within the same cell.
          
          • Stacking is only allowed if all shapes share the **same depth**.
          
          • Since the shapes can be stacked, multiple shapes may occupy the same cell. If you have any questions about stacking different shapes in one cell, feel free to ask for clarification. You do not need to remove an existing shape to stack a new one in the same cell.                                            

          • When multiple shapes are placed in a single cell:
          
              - The **first mentioned shape** is placed **at the bottom**.
              
              - Later shapes stack **on top**.
              
          • Do not stack same shapes in the same cell.
          
            Example:
            
                Correct: put(board, shape='washer', color='red', x=1, y=1)
                
                         put(board, shape='nut', color='green', x=1, y=1)
                         
                Incorrect: put(board, shape='washer', color='red', x=1, y=1)
                
                           put(board, shape='washer', color='green', x=1, y=1)                           
\\\\        
        ---
        
        IMPLEMENTATION DETAILS
        
          • When placing **multiple shapes**, use **loops** or repeated `put` calls as needed.
          
            Example:
            
                for row in [1,4]:
                
                    put(board, shape, color, row, y)
        
                for col in [2,5]:
                
                    put(board, shape, color, x, col)
        
                for row in [1,3]:
                
                    for col in [1,4]:
                    
                        put(board, shape, color, row, col)        
        ---       

\end{prompt}
\caption{Part-3 of the prompt template used for the single agent VLM in single-turn setting.}
    \label{fig:promptstructure_sagent_part3}
\end{figure*}

\begin{figure*}
  \centering
    \begin{prompt}
        Given:
        
        - prompt: Prompt with the following details:
        
            - Two images: Image1 (reference legend for shapes), Image2 (gold structure of placements)
\\\\
        Output Format:
        
        - Respond with a JSON object:
        
          Format:
          
          \{
          
            "status": "<string>",       \# e.g. "code"
            
            "details": "<string>",     \# Python code
            
          \}

          Example:

          \{
          
               "status": "code",
               
               "details": "put(board, 'washer', 'green', x=1, y=1)

               put(board, 'nut', 'blue', x=3, y=2)"
               
          \}      
        
        Make sure the JSON is valid and parsable by Python json.loads().
\\\\
Respond with the corresponding output fields, starting with the field `[[ \#\# player\_response \#\# ]]`, and then ending with the marker for `[[ \#\# completed \#\# ]]`.

\end{prompt}
\caption{Part-4 of the prompt template used for the single agent VLM in single-turn setting.}
    \label{fig:promptstructure_sagent_part4}
\end{figure*}
    
\begin{figure*}
  \centering
    \begin{prompt}
\\
Your input fields are:

1. `prompt` (str): Prompt with goal grid and current grid state

Your output fields are:

1. `player\_response` (str): A JSON object with keys 'status' and 'details' as described above

All interactions will be structured in the following way, with the appropriate values filled in.

[[ \#\# player\_response \#\# ]]

{player\_response}

[[ \#\# completed \#\# ]]
\\\\
In adhering to this structure, your objective is: 

        You are the robot in a structure reconstructing game.
\\\\
        You will be shown three images:
        
        • Image 1 is a reference legend — it shows each shape alongside its label. Use this to identify and name shapes correctly that appear in Image 2.
        
        • Image 2 is the gold truth — the target structure that must be reconstructed.

        • Image 3 shows the current state of placements on an 8×8 grid — it reflects all placements made so far. At the start of the game, the grid is empty.        
        
        • Using the labels from Image 1, interpret Image 2 accurately. Identify each placed shape, reason carefully about its grid position, occupied cells, and stacking order. Then generate action sequence to reconstruct the target structure.
        The response should be a JSON object as described in the Output Format section below.    
\\\\
The rest of the prompt about environmental constraints, API usage and the response format is same as the prompt used for single agent VLM in single turn.
\end{prompt}
\caption{Prompt template used for the single agent VLM in multi-turn.}
    \label{fig:promptstructure_sagent_mt}
\end{figure*}        
\begin{figure*}
  \centering
    \begin{prompt}
\\
Your input fields are:

1. `prompt` (str): Prompt with reference image and target image

Your output fields are:

1. `instruction` (str): Natural-language instruction for the robot

All interactions will be structured in the following way, with the appropriate values filled in.

[[ \#\# instruction \#\# ]]

{instruction}

[[ \#\# completed \#\# ]]
\\\\
In adhering to this structure, your objective is: 

        You are the human user.
        
        Your task is to instruct a robot to reconstruct the given goal structure in Image2. You will have to generate all the necessary instructions at once.
\\\\
        You will be shown two images:
        
        • Image 1 is a reference legend — it shows each shape alongside its label. Use this to identify and name shapes correctly that appear in Image 2.
        
        • Image 2 is the gold truth — the target structure that must be reconstructed.
        
        • Using the labels from Image 1, interpret Image 2 accurately. Identify each placed shape, reason carefully about its grid position, occupied cells, and stacking order. Then generate textual placement instructions that would reconstruct the object shown in Image 2 from scratch on an empty grid.
\\\\
        Grid Orientation
        
        • Rows are numbered 1 to 8, top to bottom
        
        • Columns are numbered 1 to 8, left to right
        
        • A cell is always described as row, column — strictly in that order.
        
            Do not confuse rows and columns.
            
            row:3, col:4 and row:4, col:3 are different cells and must not be confused. 
\\\\
        Shape identification
        
        • Nut: a square shape occupying exactly one cell.
        
        • bridge-h: a horizontal bar occupying two adjacent cells in the same row.
        
        • Do not interpret a bridge-h as two separate nuts placed side by side. If relevant, also distinguish vertical extensions similarly when visible.          
\\\\
        Preferred Instruction Style:
        
        • Describe shapes, regions, and patterns.
        
        • Use relative spatial language (center, corners, edges, top/bottom/left/right).
        
        • Use symmetry, repetition, and grouping where possible.
        
        • Describe what the grid should look like, and how to reconstruct it
\\\\
        Communication Strategy:
        
        • Assume the follower understands shapes and patterns.   
      
\end{prompt}
\caption{Part-1 of prompt template used for the Programmer in two-agent VLM in single-turn.}
    \label{fig:promptstructure_twoagent_prog_st_1}
\end{figure*}        

\begin{figure*}
  \centering
    \begin{prompt}
\\
        Stacking constraints:
        
        • Instructions must respect stacking dependencies.
        
        • Always describe actions from bottom to top: supporting objects must be placed before any objects that rest on them.
        
        • Do NOT instruct placement of top-level objects before their supports exist.
        
        • Violating this order may lead to incorrect reconstruction.
\\\\
        Partially occluded stacked shapes handling:
        
        • When shapes are stacked, lower shapes may be partially or fully hidden. Use these cues to identify them.
        
        • Infer the hidden shape from visible evidence:
        
            Visible corners — If only the corners of a diamond are visible, the hidden shape is likely a washer.
            
            Half of a shape extending — If part of a shape extends into the next cell horizontally, it is likely a bridge-h.
            
            Half of a shape extending — If part of a shape extends into the next cell vertically, it is likely a bridge-v.
            
            When colors match across layers, do not assume it is a single shape — interpret the stack carefully in bottom-to-top order  
\\\\            
        Rules:
        
        • Rows spread from top to bottom vertically ranged from 1 to 8; Columns spread from left to right horizontally ranged from 1 to 8
        
        • 'Nut' is a square; 'bridge-h' is a long horizontal bar, do not confuse it with Nut. 'Nut' occupies a single cell and 'bridge-h' occupies two cells.
        
        • If a region or pattern is already correct, do not mention it.

        Examples:
        
        "Create a vertical stack with a red washer below a blue screw."
        
        "Attach a yellow nut to the right side of green horizontal bridge."
        
        "Mirror the structure on the left side onto the right side."
        
        "First place the nut, then place screw on top of it."
\\\\
Inputs:

    - prompt: Prompt with the following details:
    
        - Image 1 contains reference legend for the shapes used in an object.
        
        - Image 2 contains a target object on a 8x8 grid.         
\\\\
Outputs:

    - Your goal is to generate natural language instructions to construct the target object in Image2
    
    - Do not generate anything else

Respond with the corresponding output fields, starting with the field `[[ \#\# instruction \#\# ]]`, and then ending with the marker for `[[ \#\# completed \#\# ]]`.

\end{prompt}
\caption{Part-2 of prompt template used for the Programmer in two-agent VLM in single-turn.}
    \label{fig:promptstructure_twoagent_prog_st_2}
\end{figure*}        
\begin{figure*}
  \centering
    \begin{prompt}
\\
Your input fields are:

1. `prompt` (str): Prompt with reference image, target image and Robot's current grid.

Your output fields are:

1. `instruction` (str): Natural-language instruction for the robot

All interactions will be structured in the following way, with the appropriate values filled in.

[[ \#\# instruction \#\# ]]

{instruction}

[[ \#\# completed \#\# ]]
\\\\
In adhering to this structure, your objective is: 

        You are the human user.
        
        Your task is to collaborate with a robot to reconstruct the given goal structure in Image2 by generating natural language instructions.
\\\\
        You will be shown three images:
        
        • Image 1 is a reference legend — it shows each shape alongside its label. Use this to identify and name shapes correctly that appear in Image 2.
        
        • Image 2 is the gold truth — the target structure that must be reconstructed.
        
        • Image 3 shows the current state of robot reconstruction on an 8×8 grid — it reflects all placements made so far. At the start of the game, the grid is empty.

        • Using the labels from Image 1, interpret Image 2 accurately. Identify each placed shape, reason carefully about its grid position, occupied cells, and stacking order. Then generate textual placement instructions that would reconstruct the object shown in Image 2 from scratch on an empty grid.
\\\\
        Grid Orientation
        
        • Rows are numbered 1 to 8, top to bottom
        
        • Columns are numbered 1 to 8, left to right
        
        • A cell is always described as row, column — strictly in that order.
        
            Do not confuse rows and columns.
            
            row:3, col:4 and row:4, col:3 are different cells and must not be confused. 
\\\\
        Shape identification
        
        • Nut: a square shape occupying exactly one cell.
        
        • bridge-h: a horizontal bar occupying two adjacent cells in the same row.
        
        • Do not interpret a bridge-h as two separate nuts placed side by side. If relevant, also distinguish vertical extensions similarly when visible.          
\\\\
        Preferred Instruction Style:
        
        • Describe shapes, regions, and patterns.
        
        • Use relative spatial language (center, corners, edges, top/bottom/left/right).
        
        • Use symmetry, repetition, and grouping where possible.
        
        • Describe what the grid should look like, and how to reconstruct it
\\\\
        Communication Strategy:

        • Focus on completing one meaningful pattern or region at a time.
        
        • Assume the follower understands shapes and patterns.   
      
\end{prompt}
\caption{Part-1 of prompt template used for the Programmer in two-agent VLM in multi-turn.}
    \label{fig:promptstructure_twoagent_prog_mt_1}
\end{figure*}        

\begin{figure*}
  \centering
    \begin{prompt}
\\
        Stacking constraints:
        
        • Instructions must respect stacking dependencies.
        
        • Always describe actions from bottom to top: supporting objects must be placed before any objects that rest on them.
        
        • Do NOT instruct placement of top-level objects before their supports exist.
        
        • Violating this order may lead to incorrect reconstruction.
\\\\
        Partially occluded stacked shapes handling:
        
        • When shapes are stacked, lower shapes may be partially or fully hidden. Use these cues to identify them.
        
        • Infer the hidden shape from visible evidence:
        
            Visible corners — If only the corners of a diamond are visible, the hidden shape is likely a washer.
            
            Half of a shape extending — If part of a shape extends into the next cell horizontally, it is likely a bridge-h.
            
            Half of a shape extending — If part of a shape extends into the next cell vertically, it is likely a bridge-v.
            
            When colors match across layers, do not assume it is a single shape — interpret the stack carefully in bottom-to-top order  
\\\\            
        Rules:
        
        • Rows spread from top to bottom vertically ranged from 1 to 8; Columns spread from left to right horizontally ranged from 1 to 8
        
        • 'Nut' is a square; 'bridge-h' is a long horizontal bar, do not confuse it with Nut. 'Nut' occupies a single cell and 'bridge-h' occupies two cells.
        
        • Use the robot grid to identify which high-level patterns or regions are incomplete.
        
        • If a region or pattern is already correct, do not mention it.
        
        • Generates the instructions to build incorrect or missing elements.

        • If clarification input is non-empty, address it directly and clearly.

        • If the entire grid matches the goal grid and reconstruction\_status is True, reply only with the word "DONE".

        • Otherwise, generate instructions that advances the reconstruction

        • Always output either an instruction or "DONE" — nothing else.
        
        Examples:
        
        "Create a vertical stack with a red washer below a blue screw."
        
        "Attach a yellow nut to the right side of green horizontal bridge."
        
        "Mirror the structure on the left side onto the right side."
        
        "First place the nut, then place screw on top of it."
\\\\
Inputs:

    - prompt: Prompt with the following details:
    
        - Image 1 contains reference legend for the shapes used in an object.
        
        - Image 2 contains a target object on a 8x8 grid.  

        - Image 3 contains the current grid status filled by the robot; At the beginning of the game, the grid would be empty.

        - clarification: Optional question received from the robot; may be empty.

        - acknowledgement: Optional message received from the robot; may be empty.        
\\\\
Outputs:

    - Your goal is to generate natural language instructions to construct the target object in Image2
    
    - Do not generate anything else

Respond with the corresponding output fields, starting with the field `[[ \#\# instruction \#\# ]]`, and then ending with the marker for `[[ \#\# completed \#\# ]]`.

\end{prompt}
\caption{Part-2 of prompt template used for the Programmer in two-agent in multi-turn.}
    \label{fig:promptstructure_twoagent_prog_mt_2}
\end{figure*}        
\begin{figure*}
  \centering
    \begin{prompt}
\\

Your input fields are:

1. `prompt` (str): Prompt with Instruction from User and current grid state

Your output fields are:

1. `player\_response` (str): A JSON object with keys 'status' and 'details' as described above

All interactions will be structured in the following way, with the appropriate values filled in.

[[ \#\# player\_response \#\# ]]

{player\_response}

[[ \#\# completed \#\# ]]
\\\\
In adhering to this structure, your objective is: 

        You are the robot (the instruction follower) in a collaborative grid-filling game.
\\\\
        You will be shown three images:
        
        • Image 1 is a reference legend — it shows each shape alongside its label. Use this to identify and name shapes correctly that appear in Image 2.
        
        • Image 2 shows the current state of placements on an 8×8 grid — it reflects all placements made so far. At the start of the game, the grid is empty.
        
        • Using the labels from Image 1, interpret Image 2 accurately. Identify each placed shape, reason carefully about its grid position, occupied cells, and stacking order. Then respond with a JSON object as described in the Output Format section below.   
\\\\
The rest of the prompt about environmental constraints and API usage is same as the prompt used for single agent VLM in single turn. The response format is changed and is shown below.

        Given:

        - prompt: Prompt with the following details:
            - Two images: Image1 (reference legend for shapes), Image2 (current state of placements)

            - The instruction from the user describing what to do
        
        Output Format:

        - Respond with a JSON object:

          Format:

          {

            "status": "<string>",       \# e.g. "clarification" or "code"

            "details": "<string>",     \# Python code when status="code", or plain text when status="clarification"

          }
        
          If clarification needed:

          {

            "status": "clarification",

            "details": "Which color washer should I use?"

          }

          If providing acknowledgement:

          {

            "status": "acknowledgement",

            "details": "Yes, I know"

          }           
        
          If executing code:

          {

               "status": "code",

               "details": "put(board, 'washer', 'green', x=1, y=1)
               
               put(board, 'nut', 'blue', x=3, y=2)"

          }      
        
        Make sure the JSON is valid and parsable by Python json.loads().
Respond with the corresponding output fields, starting with the field `[[ \#\# player\_response \#\# ]]`, and then ending with the marker for `[[ \#\# completed \#\# ]]`.

\end{prompt}
\caption{Prompt template used for the Robot in two-agent VLM in multi-turn.}
    \label{fig:promptstructure_twoagent_robot_mt}
\end{figure*}        

\begin{figure*}
  \centering
    \begin{prompt}
\\
Goal:Grid levels (bottom to top):

Level 1:

row: 3, col: 1: {'shapes': ['washer'], 'colors': ['red']}

row: 3, col: 2: {'shapes': ['washer'], 'colors': ['red']}

Level 2:

row: 3, col: 1: {'shapes': ['bridge-h'], 'colors': ['green']}

Level 3:

row: 3, col: 1: {'shapes': ['nut'], 'colors': ['yellow']}
\\\\
Difference grid: Differences between target grid and player's grid (bottom to top):

Level 1:

row: 3, col: 1: Identical

row: 3, col: 2: Identical

Level 2:

row: 3, col: 1: Identical

row: 3, col: 2: Extra {'shapes': ['screw'], 'colors': ['blue']}

Level 3:

row: 3, col: 1: Missing {'shapes': ['nut'], 'colors': ['yellow']}

\end{prompt}
\caption{Example difference grid: Textual information for the Programmer on where the Robot's current grid matches and differs}
    \label{fig:promptstructure_difference_grid}
\end{figure*}   

\end{document}